\documentclass{article}
\usepackage{spconf}
\usepackage{times}
\usepackage{epsfig}
\usepackage{graphicx}
\usepackage{amsmath}
\usepackage{amssymb}
\usepackage{booktabs}
\usepackage{tabularx}
\usepackage{textcomp}
\usepackage{soul}
\usepackage{pdfpages}
\usepackage{import}
\usepackage{multirow}

\usepackage{verbatim}
\usepackage[export]{adjustbox}
\usepackage{caption}
\usepackage[colorlinks = true,
            allcolors=.,
            urlcolor=blue,]{hyperref}

\title{\textit{LiFi}: Towards Linguistically Informed Frame Interpolation}
\name{
\parbox{\linewidth}{\centering
Aradhya Neeraj Mathur$^{1,*}$\thanks{*Equal Contribution}, Devansh Batra$^{1,*}$, Yaman Kumar$^{1,2*}$, \linebreak {Rajiv Ratn Shah$^{1}$,  Roger Zimmermann$^{3}$}}}
\address{
\textsuperscript{1}IIIT-Delhi, 
\textsuperscript{2}Adobe, \textsuperscript{3}NUS\\
\texttt{\textsuperscript{1}\{aradhyam, yamank, rajivratn\}@iiitd.ac.in}, \texttt{\textsuperscript{1}devansh.batra@midas.center},\\ \texttt{\textsuperscript{2}ykumar@adobe.com, \textsuperscript{3}rogerz@comp.nus.edu.sg}}

\begin{document}
\maketitle

\begin{abstract}
\begin{quote}
	In this work, we explore a new problem of frame interpolation for speech videos. Such content today forms the major form of online communication. %In the recent times, most face to face communication has moved online but frame drops continues to plague video calls. In this paper 
	We try to solve this problem by using several deep learning video generation algorithms to generate the missing frames. We provide examples where computer vision models despite showing high performance on conventional non-linguistic metrics fail to accurately produce faithful interpolation of speech videos. With this motivation, we provide a new set of linguistically-informed metrics specifically targeted to the problem of speech videos interpolation. We also release several datasets to test computer vision video generation models of their speech understanding.
\end{quote}
\end{abstract}
%%%%%%%%% BODY TEXT
\vspace{-1em}
%%%%%%%%%%%%%%%%%%%%%%%%%%%%%%%%%%%%%%%%%%%%%%%%%%%
\section{Introduction} \vspace{-1em}
Video frame interpolation and extrapolation is the task of synthesizing new video frames conditioned on the context of a given video~\cite{liu-et-al}. Contemporary applications of such interpolation are video playback software for increasing frame rates, video editing software for creating slow motion effects, and virtual reality software to decrease resource usage. From the early per-pixel phase-based shifting approaches such as by \cite{Meyer15Phase}, current approaches have shifted to generating video frames using techniques like optical flow or stereo methods~\cite{liu-et-al, jiang2018super}. These types of approaches typically involve two steps: motion estimation and pixel synthesis. While motion estimation is needed to understand the movement of different objects of the images across frames, pixel synthesis focuses on the generation of the new data.\\
%\noindent

A related task to video frame interpolation is talking face generation. Here, given an audio waveform, the task is to automatically synthesize a talking face \cite{kumar2017obamanet,chen2018lip}. In recent times, these approaches have become popular with both academic and non-academic purposes \cite{Guardian-fake-news}. While, on one hand, they are being used to extend speechreading models to low resource languages \cite{kumar2019harnessing}%and to model the geometry of humans in different dresses \cite{pumarola20193dpeople}
, on the other, many of them are also used to generate fake news and paid content as well.

The literature on talking face generation, as well as video interpolation and extrapolation use the conventional video quality metrics such as root mean squared (RMS) distance, structural similarity index, \textit{etc.} to measure the quality of generated videos \cite{wang2004image}. Although this evaluates the quality of generated pixels well, but none of the research works show how it translates to linguistically plausible video frames. This is important since speech being a linguistic problem,f cannot be solely addressed by non-linguistic metrics like RMS distance. The goal of this paper is to investigate how the different video interpolation and extrapolation algorithms are able to capture linguistic differences between the generated videos. 

Speech as a natural signal is composed of three parts \cite{massaro2014speech}: visual modality, audio modality and the context in which it was spoken (crudely the role played by language). Correspondingly, there are three tasks for modeling speech: speech-reading (or popularly known as lipreading) \cite{kumar2019harnessing,chung2016lip}, speech recognition (or ASR) \cite{yu2016automatic} and language modeling \cite{mikolov2013distributed}. The part of speech which is closest to the speech video generation task is the visual modality of speech; and visemes are the fundamental units of this part of speech. Calculating metrics such as mean squared error (MSE) over the whole video does not directly yield any information on these aspects of speech which makes us question the faithfulness of the thus attained reconstruction. Therefore, the focus of this work is to investigate video generation model's understanding of the visual speech modality. To this end, we propose six tasks for looking at different aspects of the visual speech modality. 
Hence, with this work, we try to make the following contributions:
% \begin{enumerate}
%\noindent
 1. We explore different video interpolation and extrapolation networks for usage on speechreading videos and propose a new auxiliary method of ROI loss using an visemic ROI unit for attaining faithful reconstructions.\\
 2. For the first time in literature, we test the video frame generation algorithms on different aspects of speech such as visemic completion, generating prefix and suffix from context, word-level understanding, \textit{etc}. These facets of language are critical to a model's language understanding. We show that most of these networks are not able to capture the language aspect of a speech video.\\
 3. We release six new challenge datasets corresponding to different language aspects. These have been verified automatically and manually and are meant to facilitate reproduction and follow-up testing and interpretation.

%%%%%%%%%%%%%%%%%%%%%%%%%%%%%%%%%%%%%%%%%%%%%%%%%%%%%%%%%%%%5

\begin{table*}[htp]
\begin{minipage}[b]{0.50\textwidth}
    
	\centering
%\begin{table*}[!htp]
%\centering
	\begin{tabular}{|c|c|c|c|c|}
        \hline
		Viseme & FCN3D & LSTM & FCN3D & Super \\
		& & & (ROI) & SloMo \\
		\hline
		\hline
% 		\\
		\multicolumn{5}{|c|}{\textbf{Viseme Corruption}} \\
		\hline
% 		\textbf{FCN3D} & 75 & 0.03294822413 & 0.1267357732 & 0.5301524016 & 14.83449745 \\
		\textbf{@} & 0.8974 & 0.7489 & 0.9086 & 0.9660 \\
		\hline
		\textbf{a} & 0.8838 & 0.7410 & 0.9152 & 0.9566 \\
		\hline
		\textbf{E} & 0.8902 & 0.7425 & 0.9231 & 0.9656 \\
		\hline
		\textbf{f} & 0.8922 & 0.7465 & 0.9197 & 0.9617 \\
		\hline
		\textbf{i} & 0.8923 & 0.7387 & 0.9239 & 0.9761 \\
		\hline
		\textbf{k} & 0.8910 & 0.7434 & 0.9242 & 0.9684 \\
		\hline
		\textbf{O} & 0.9059 & 0.7513 & 0.9213 & 0.9558 \\
		\hline
		\textbf{p} & 0.8921 & 0.7415 & 0.9256 & 0.9715 \\
		\hline
		\textbf{r} & 0.8998 & 0.7446 & 0.9255 & 0.9731 \\
		\hline
		\textbf{s} & 0.8920 & 0.7497 & 0.9238 & 0.9662 \\
		\hline
		\textbf{S} & 0.8829 & 0.7398 & 0.9237 & 0.9556 \\
		\hline
		\textbf{t} & 0.8945 & 0.7443 & 0.9268 & 0.9702 \\
		\hline
		\textbf{T} & 0.9087 & 0.7436 & 0.9185 & 0.9771 \\
		\hline
		\textbf{u} & 0.8918 & 0.7503 & 0.8936 & 0.9395 \\
		\hline
		% Model & Corruption (\%) & MSE & L1 & SSIM & PSNR \\
 % \\		
		
	\end{tabular}
	\caption{SSIM obtained upon reconstruction of corrupted videos in the Visemic Corruption dataset, averaged across the videos being corrupted for each viseme.}
	\label{tab:metric_table_viseme_full}
%\end{table*}
\end{minipage}\qquad
\begin{minipage}[b]{0.50\textwidth}
%\begin{table*}[htp]
%\begin{minipage}[b]{0.60\textwidth}
    % \scriptsize
	\centering
	\scalebox{0.85}{
	\begin{tabular}{|c|c|c|c|c|}
		\hline
% 		\\
		Model & SSIM & PSNR & SSIM & PSNR \\
		\hline
		\textbf{Corruption} & \multicolumn{2}{|c|} {40\%} & \multicolumn{2}{|c|} {75\%} \\
		\hline
		\hline
% 		\\
		\multicolumn{5}{|c|}{\textbf{Random Corruption}} \\
		\hline
		
		\textbf{FCN3D} & 0.9271 & 23.7161 & 0.8554 & 21.8354 \\
		\hline
		\textbf{BDLSTM} & 0.8119 & 21.6103 & 0.8199 & 22.3904 \\
		\hline
		\textbf{FCN3D + ROI} & \textbf{0.9326} & \textbf{28.3123} & \textbf{0.8654} & \textbf{
		24.7521}\\
		\hline 
		\textbf{Super SloMo} & \textbf{0.9849} & \textbf{30.3459} & \textbf{0.9603} & \textbf{28.2660}\\
		\hline
		\hline
% 		\\
		\multicolumn{5}{|c|}{\textbf{Prefix Corruption}} \\
% 		\hline
		% Model & Corruption (\%) & MSE & L1 & SSIM & PSNR \\
		\hline
		\textbf{FCN3D} & 0.7680 & 18.4513 & 0.5230 & 14.7208 \\
		\hline
		\textbf{BDLSTM} & \textbf{0.8119} & \textbf{21.6103} & \textbf{0.6288} & \textbf{17.6097} \\
		\hline
		\textbf{FCN3D + ROI} & 0.7721 & 19.6718 & 0.5208 & 15.3935\\
		\hline
		\textbf{Super SloMo} & \textbf{0.8411} & \textbf{23.1864} & \textbf{0.7387} & \textbf{20.8218}\\
		\hline
		\hline
% 		\\
		\multicolumn{5}{|c|}{\textbf{Suffix Corruption}} \\
		\hline
		% Model & Corruption (\%) & MSE & L1 & SSIM & PSNR \\
		\textbf{FCN3D} & 0.7816 & 18.7518 & 0.5301 & 14.8344 \\
		\hline
		\textbf{BDLSTM} & \textbf{0.7682} & \textbf{20.4863} & \textbf{0.6281} & \textbf{17.8500} \\
		\hline
		\textbf{FCN3D + ROI} & 0.7774 & 19.7790 & 0.5161 & 15.3414\\
		\hline
		
		\textbf{Super SloMo} & \textbf{0.8140} & \textbf{22.4334} & \textbf{0.6952} & \textbf{19.8319}\\
		\hline

	\end{tabular}}
	\caption{Metrics obtained on the test set for the different experiments conducted. For each experiment, two different levels of corruption were tested, a low 40\% and a high 75\% corruption. It also highlights how the behaviors of models vary across different experiments, thus further reaffirming the need for a more comprehensive test suite. We share more detailed metrics in appendix Table \ref{tab:appendix_metric_table}. }
	\label{tab:metric_table1}
% \end{table*}

\end{minipage}\qquad
\end{table*}

%%%%%%%%%%%%%%%%%%%%%%%%%%%%%%%%%%%%%%%%%%%%%%%%%%%%%%%%%%%%5

\section{Evaluation Methods}\vspace{-1em}
\label{sec:evaluation methods}

\begin{figure}[!hb]

 \centering
 \begin{tabular}{c}
 \includegraphics[scale=0.22]{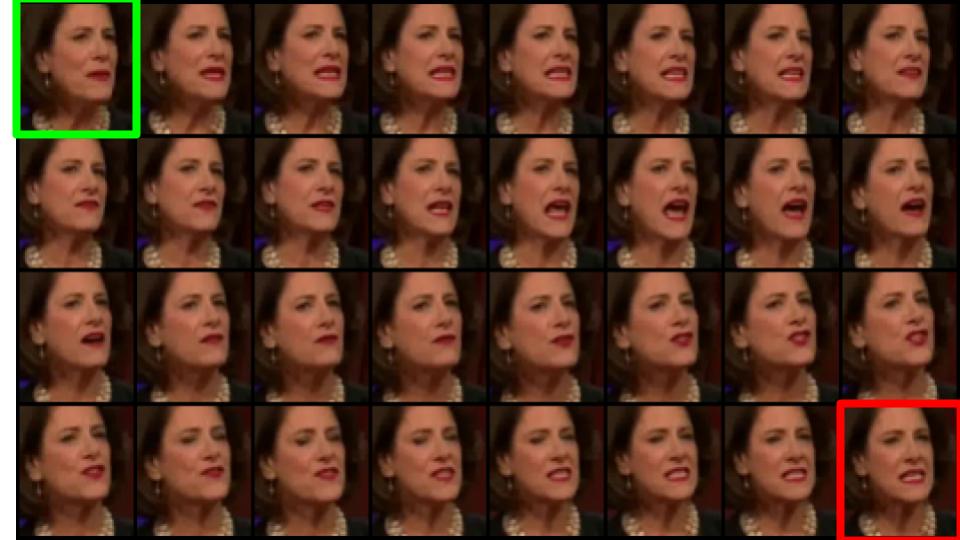}\\
 \includegraphics[scale=0.22]{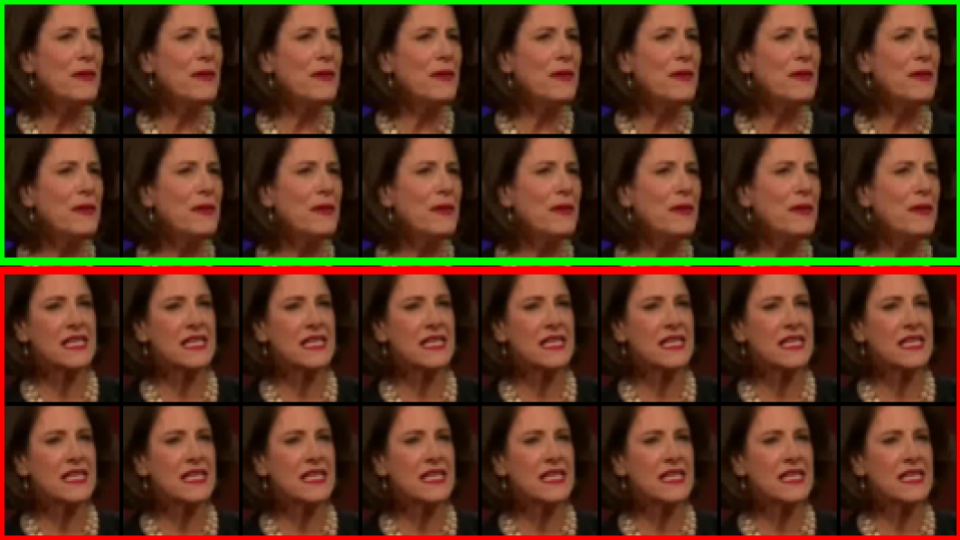} \\ 
 \textbf{MSE}:  0.0093898 \;% MSE: \; 0.01086 \\
 \textbf{PSNR}:  20.273426 \;% PSNR: \; 19.64074 \\
 \textbf{SSIM}:  0.6174879 % SSIM: \; 0.7619643
 \end{tabular}

 \caption{To show the ineffectiveness of the existing metrics such as PSNR, SSIM, MSE we create a new synthetic video clip of size 32 frames by replicating the first frame 16 times and the last frame 16 times in the second image. We notice that even though contextually the information is incorrect since only two frames are used to construct the complete video, the metrics show high values indicating that the synthetic video is a faithful reconstruction of the original video. This result reinforces the fact that we need better metrics to compensate for the underlying context and comes in par with the metrics obtained by training FCN3D in Table \ref{tab:metric_table1}. However, the proposed method takes into account the visemic reconstruction thus enabling to get a more faithful reconstruction. }
 
 \label{fig:frame_diff}
\end{figure}

\begin{figure*}[!ht]

 \centering
 \begin{tabular}{c}
 \includegraphics[scale=0.18]{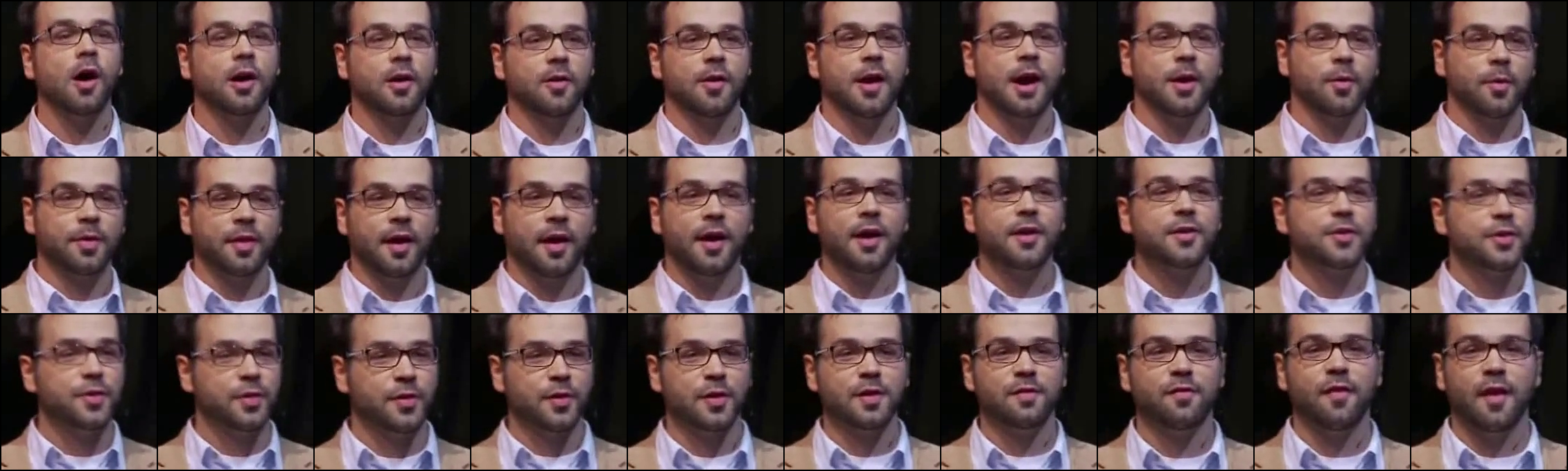}\\
 \includegraphics[scale=0.18]{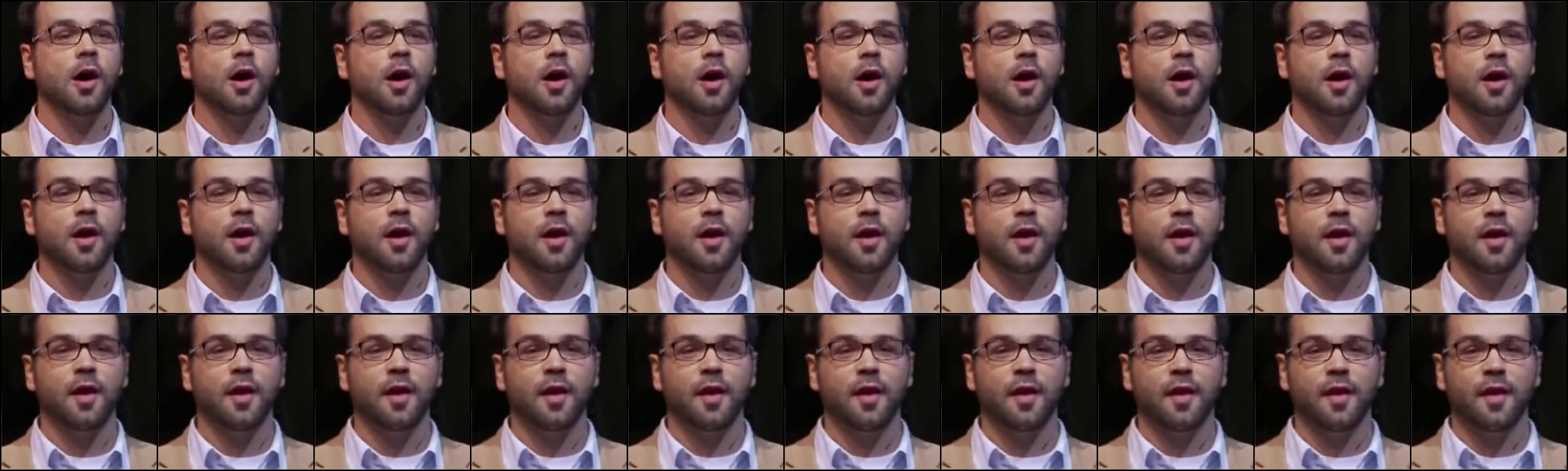} \\ 
 \textbf{MSE}:  0.0088 \;% MSE: \; 0.01086 \\
 \textbf{PSNR}:  20.5415 \;% PSNR: \; 19.64074 \\
 \textbf{SSIM}:  0.9521 % SSIM: \; 0.7619643    
 \end{tabular}

 \caption{The above images are part of the original and Super-SloMo interpolated videos for the phrase `I don't exactly walk around with a hundred and thirty five million dollars in my wallet'. Although the interpolated frames have an excellent quality, and achieve very high scores on the metrics, zooming on them reveals that they are not able to reproduce the lip movements. For instance, in the figure above, the speaker closes his mouth several times but in the bottom one, the speaker does not close his mouth at all. This interpolated video despite achieving excellent scores, will look asynchronous to a viewer.}
 
 \label{fig:frame_diff_superSlomo}
\end{figure*}

Most of the previous research on speech video reconstruction and interpolation focuses on conventional metrics like MSE, SSIM and PSNR, which although ensure good quality reconstructions \cite{wolterink2017deep,jiang2018super}, can be misleading in terms of the nuances of the underlying dataset. Videos of the same person saying two different things can have fairly high values of the mentioned metrics thus indicating faithful reconstruction, which  makes the evaluation difficult (Figures~\ref{fig:frame_diff}, \ref{fig:frame_diff_superSlomo})\footnote{We provide two examples in the Figures~\ref{fig:frame_diff} and \ref{fig:frame_diff_superSlomo} and a few more in the Figure \ref{fig:appendix_frame_diff} in the Appendix.}. We propose the following evaluation methods and corresponding datasets~\footnote{We make the datasets available at  \url{https://sites.google.com/view/yaman-kumar/speech-and-language/linguistically-informed-lrs-3-dataset}. Training scripts for the models in Section~\ref{subsec:models} are available at \url{https://github.com/midas-research/linguistically-informed-frame-interpolation/}} (Table~\ref{tab:metrics for dataset released}) for the speech videos to take into account the underlying language and speech information.\\

% \vspace{-1em}
\begin{table}[htbp]
    % \scriptsize
	\centering
	\footnotesize
\scalebox{0.85}{
	\begin{tabular}{|l|l|l|l|l|}
		\hline 
		\textbf{Corruption Types} & \textbf{\# Sp} & \textbf{\# Ut} & \textbf{\# WI} & \textbf{Vo}\\ \hline
		
		\textbf{\textsc{RandomFrame and}} & 4,004 & 31,982 & 356,940 & 17,545\\
		\textbf{\textsc{Extreme Sparsity}} & & & &\\
		\hline 
% 		\textbf{\textsc{Extreme Sparsity}} &  XXX & XXX & XXX & XXX & XXX \\ \hline 
		\textbf{\textsc{Visemic}} &  2,883 & 6,152 & 338,207 & 16,663\\ \hline
		
	\textbf{\textsc{InterWord}} &  3,008 & 10,421 & 141,850 & 12,621 \\ \hline
	\textbf{\textsc{IntraWord}} &   2,756 & 10,360 & 141,296 & 11,824\\ \hline
	\textbf{\textsc{Prefix and Suffix}} &  4,004 & 31,982 & 356,940 & 17,545\\ \hline 
	%\color{green}
	\textbf{\textsc{POS Tagged Corruptions}} & 4,004 & 31,982 & 356,940 & 17,545\\ \hline
% 	\textbf{\textsc{Suffix}} &  XXX & XXX & XXX & XXX & XXX  \\ \hline
	\end{tabular}}
	\caption{
		\small
		\label{tab:metrics for dataset released} Statistics for the challenge datasets made for each task. Legend: \#~Sp:~Number of speakers, \#~Ut:~Number of utterances, \#~WI:~Number of word instances, Vo: Vocabulary. 
% 		AVL:~Average Video Length in seconds
		}
\vspace{-1em}
\end{table}

%%\noindent
\vspace{-1em}

\textbf{1. \textsc{RandomFrame:}}
In this type of frame corruption, the frames at random timestamps are replaced with white noise. In a real scenario, the missing frames are dropped however the indices of the dropped frames are generally available \cite{rowe1994mpeg}. To indicate the missing frames, we insert white noise images in their place. We do this type of corruption in two settings: 40\% and 75\% corruption rate. We call the 75\% case as one with the \textit{extreme sparsity}. A model that can work in such conditions can deal with extremely sparse speech videos and high dropping rates. Random corruption tests if the model effectively interpolates between different poses, facial expressions and probable word completions.\\
% Eg: XXX
%\noindent
%\vspace{-1em}

\textbf{2. \textsc{Visemic Corruption:}}
A viseme is a specific facial image corresponding to a particular sound \cite{massaro2014speech}. In this type of corruption, we choose visemically equivalent sub-words in two or more words and corrupt the first viseme within the subword. For example, consider the words `million' and `billion', both are visemically equivalent. We corrupt the first common viseme - `p' and show the results for such corruption in Table~\ref{tab:metric_table2}. The dataset also includes the cases where the visemic equivalence does not occur at the start of the word. For example, the first common viseme which is corrupted is `@', in `Probably' and `Possibly', when the phonetic phrase ``ebli" is spoken in both. The timestamps of specific phonemes are obtained through the Montreal Forced Aligner~\cite{mcauliffe2017montreal}. After that, the phonemes were converted to visemes. Through this type of corruption, we test the model's knowledge of specific visemes given the context. We list the SSIM metrics after reconstructing each viseme is detailed in Table~\ref{tab:metric_table_viseme_full} of the Appendix.\\
%\noindent
%\vspace{-1em}

\textbf{3. \textsc{InterWord and IntraWord Corruptions:}}
In this type of corruption, we corrupt 80\% of approximately 20,000 unigram and bigram combinations. Corruption was done such that for a particular gram, 10\% each of the prefix and suffix of the overall gram remains in the original state after the corruption. For example, consider the bigram `United States', the corruption of this bigram results in `Un*********es'. A model that works well on such corruptions understands the context of the words well and is robust to corruptions during the transition from one word to another. \\
%\noindent
%\vspace{-1em}

\textbf{4. \textsc{Prefix Completion:}}
For this corruption, we remove the start frames of a video. We do this for two levels of corruptions- a low level, 40\% and a high level corruption (75\%).\\
%\vspace{-1em}

\textbf{5. \textsc{Suffix Completion:}}
Similar to the \textsc{Prefix Completion} test, we corrupt the ending frames from the sequence of input, which is also performed for 40\% and 75\% corruption. These two tests make sure that the model understands both the word start and word end and is not biased towards either. While predicting word start requires context from the previous word and the remaining part of the word, predicting the end of the word requires context from the prefix and the next word. \\

\textbf{6. \textsc{POS Tagged Corruptions:}}
{
%\color{green}
In this type of corruption, we identify nouns, pronouns, verbs, adjectives and determinants using Spacy's English language POS tagger\footnote{\url{https://spacy.io/api/tagger/}}. 
The corresponding timestamps for all such tagged words were obtained through MFA. Following this, we decided to corrupt these words for further study. For example, in a video from LRS3, the speaker says \textit{`The short answer to the question is that no, it's not the same thing'}. The POS tagger identifies the three words \textit{`answer', `question'} and \textit{`thing'} as nouns. Thus all of the frames belonging to these three words were corrupted while inducing corruptions to nouns.

By introducing such corruptions, we test how well do the models understand semantic features of the language spoken. For example, for a particular model proper nouns may be more difficult to predict than pronouns. This test helps to bring such nuances into forefront.
}
\vspace{-1em}
\section{Approach}\vspace{-1em}
\label{sec:approach}

\begin{table}[htbp]
	%\color{green}
	\centering
	\footnotesize
	\begin{tabular}{|l|l|l|l|}
		\hline \textbf{Metrics} & \textbf{Pretrain} & \textbf{Trainval} & \textbf{Test}  \\ \hline
		
		\textbf{\# Speakers} & 5091 & 4005 & 413 \\ \hline
		
		\textbf{\# Utterances} & 118,516 & 31,982 & 1321 \\ \hline
		
		\textbf{\# Word instances} & 3,894,800 & 807,375 & 9890 \\ \hline
		
		\textbf{Vocab} & 51,211 & 17,546 & 2002 \\ \hline
		
	\end{tabular}
	\caption{
	    %\color{green}
		\small
		\label{table:lrs3_overview} Overview of LRS-3 dataset used for experimentation. To train our models, we used the `Pretrain' section of the dataset and the `trainval' section as the testing set. We also sample the linguistically aware evaluation datasets (Section~\ref{sec:evaluation methods}) from the `trainval' section itself. The `test' set of LRS3 has very few videos (1321) and was dropped to simplify the distribution.}
\end{table}

\subsection{Dataset}
\label{subsec:dataset}
%\vspace{-1em}

%\noindent
We choose LRS3 dataset \cite{afouras2018lrs3} for testing out the speech video frame generation models and making the six challenge sets as described in the last section. We chose LRS3 for this task since it has a high variety of speakers (5091), number of utterances (118k) and has longer average speech length. %s(Table \ref{tab:LRS3 metrics} in Appendix) 
Other datasets are generally much less diverse both in terms of speakers and number and length of videos \cite{anina2015ouluvs2,yang2019lrw}. In LRS3, the standard frame rates are 24, 30 and 60 fps. Therefore, we choose a standard frame rate of 32 to train our models and pad it with random noise. We train and evaluate our models using the LRS3 dataset with slight modifications, as explained in Table~\ref{table:lrs3_overview}. We evaluate the models' generated videos on MSE, SSIM and PSNR metrics~\cite{jiang2018super}. 

\begin{figure*}[!ht]
\scalebox{0.95}{
    
	\centering

	\begin{tabular}{c c c}
	 Corrupt Input & Ground Truth & Generated\\
	 \includegraphics[scale=0.3]{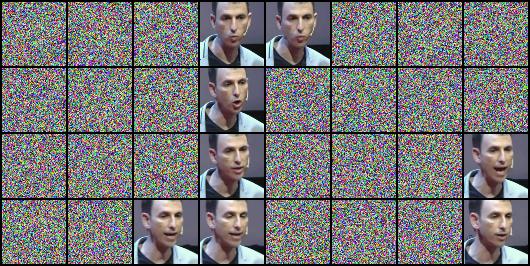} &
	 \includegraphics[scale=0.3]{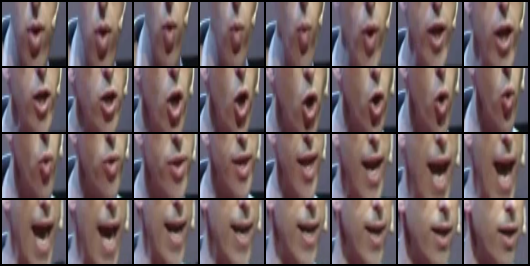}& 
		\includegraphics[scale=0.3]{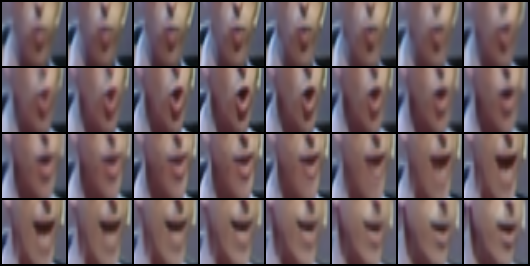} \\ 
		&
		\includegraphics[scale=0.3]{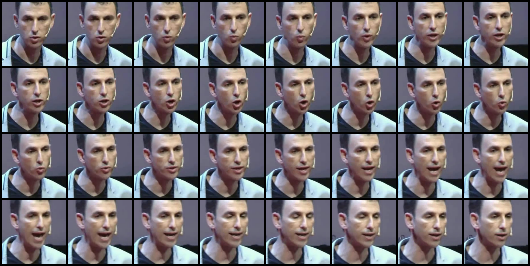}&
		\includegraphics[scale=0.3]{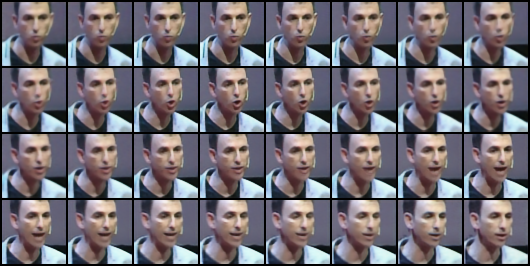}\\
		
	\end{tabular}}
	\caption{Sample Completion using ROI stream, the top image is the corrupt input, the left column contains the ground truth and the right column contains the generated output ROI and frame reconstructions.}
	\label{fig:roi_sample}
\end{figure*}

\vspace{-1em}
\begin{table}[htp]
	\centering
	\scalebox{0.9}{
	\begin{tabular}{|c|c|c|}
		\hline
% 		\\
		Model & SSIM & PSNR \\
		\hline
		\hline
% 		\\
		\multicolumn{3}{|c|}{\textbf{Viseme Corruption for Viseme}} \\
		\hline
		
		\textbf{FCN3D} & 0.8385 & 18.5201 \\
		\hline
		\textbf{BDLSTM} & 0.7269 & 17.2827 \\
		\hline
		\textbf{FCN3D + ROI} & \textbf{0.9106} & \textbf{26.5401} \\
		\hline 
		\textbf{Super SloMo} & \textbf{0.9129} & \textbf{25.2605} \\
		\hline
		\hline
% 		\\
		\multicolumn{3}{|c|}{\textbf{Word Level Corruption}} \\
% 		\hline
		% Model & Corruption (\%) & MSE & L1 & SSIM & PSNR \\
		\hline
		\textbf{FCN3D} & 0.8109 & 18.4554 \\
		\hline
		\textbf{BDLSTM} & 0.7411 & 18.066 \\
		\hline
		\textbf{FCN3D + ROI} & \textbf{0.8804} & \textbf{25.3279}\\
		\hline
		\textbf{Super SloMo} & \textbf{0.8800} & \textbf{24.5971} \\
		\hline
		\hline
% 		\\
		\multicolumn{3}{|c|}{\textbf{Multi Word Corruption}} \\
		\hline
		% Model & Corruption (\%) & MSE & L1 & SSIM & PSNR \\
		\textbf{FCN3D} & 0.8064 & 19.6111 \\
		\hline
		\textbf{BDLSTM} & 0.7411 & 18.0665 \\
		\hline
		\textbf{FCN3D + ROI} & \textbf{0.8308} & \textbf{23.1354}\\
		\hline
		
		\textbf{Super SloMo} & \textbf{0.8481} & \textbf{23.8295}\\
		\hline
		% Model & Corruption (\%) & MSE & L1 & SSIM & PSNR \\
 % \\		
		
	\end{tabular}
	}
	\caption{Metrics obtained on the newly introduced challenge datasets. It can be seen that even the simpler models trained with ROI unit embedding give results close to the baseline (Super SloMo).More details can be found in Table \ref{tab:metric_table_viseme_full} in appendix.}
	\label{tab:metric_table2}
	
\end{table}
%%%%%%%%%%%%%%%%%%%%%%%%%%%%%%%
\subsection{Models} %\vspace{-1em}
\label{subsec:models}
We perform the tests on 4 different models. We perform the baseline tests on standard methods comprising of Convolutional Bi-Directional LSTM~(BDLSTM) and Fully Convolutional Neural Network consisting of 3D convolutions~(FCN3D). Further, we compare them with the proposed 3D Fully Convolutional Network with ROI unit. FCN3D have been heavily used in 3D image synthesis based tasks such as \cite{wolterink2017deep} while convolutional BDLSTM has been used for sequential image synthesis \cite{xiang2020zooming}. For the FCN3D model, we select a 3D Denoising auto-encoder based approach which has been known to be robust for video and 3D image synthesis related tasks \cite{you2018structurally, guibas2017synthetic}. We also compare the results with NVIDIAs Super-Slo-Mo model~\cite{jiang2018super} which is a state of the art method for interpolating frames with suitable modifications for the test suite.

\begin{figure}[!htp]

	\includegraphics[width=.5\textwidth,left]{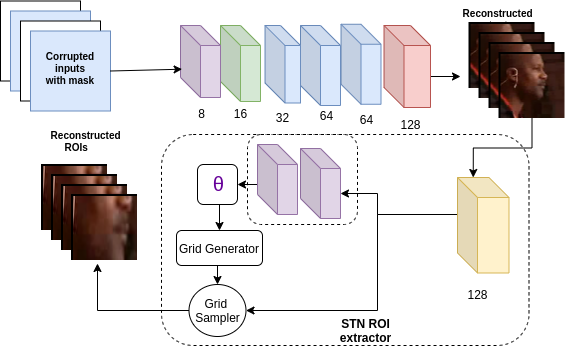}
	\caption{ The architecture of the proposed ROI loss network. The generated frames are passed through the secondary stream which learns to extract the mouth ROI. This is aimed at making the network focus more on the visemic information that each frame carries and learn the reconstruction of the ROI.}
	\label{fig:roi_arch}
\end{figure}

In this approach, we take a set of noisy frames as input and generate the denoised frames and train the network as a denoising autoencoder \cite{vincent2008extracting}. The network is expected to learn interpolations based on the available frames in the video sequence without any additional information. We test the model on two different percentages of corruptions and show the results for sparse reconstruction in the Figure~\ref{fig:roi_sample}.
The network is trained with the corrupted frames with the original frames as the output. We tried for two kinds of losses, L1 and L2, since L2 is well known to produce blurring we define the reconstruction loss in equation \ref{equation1}. 

\begin{equation}
    \label{equation1}
    \mathcal{L}_{frame} = \frac{1}{N} \sum_{x\in X}||f(x_{corr}) - y||_1
\end{equation}

% 	$ \mathcal{L}_{frame} = \frac{1}{N} \sum_{x\in X}||f(x_{corr}) - y||_1$ 
%\noindent
where $x_{corr}$ is the corrupted input image for an image $x$ in the set of images $X$. The network learns to interpolate the missing frames based on the available set of uncorrupted frames.
\vspace{-1em}
%%%%%%%%%%%%%%%%%%%%%%%%%%%%%%%%%%55

\subsubsection{Region of Interest (ROI) Unit}
%\vspace{-1 em}
We noticed that all the models described before and their losses pay attention to images as a whole, where as, the speech videos have an important component in the form of visemes. Since there are no viseme centric losses in the conventional approaches, thus they largely are governed by other aspects of the images and not the visemes. To counter this problem, we propose an ROI loss which is calculated by computing the reconstruction loss computed between the ROI, \textit{i.e.}, the mouth region of the reconstructed output and ground truth. The mouth region is extracted using the facial landmarks obtained by the Blaze package \cite{48501}. %cite

From the facial landmarks, we extract the mouth region as a proxy for viseme for a particular frame and then compute the L1 loss between the mouth region extracted for the ground truth and the generated videos. This loss helps us understand the discrepancy that might arise when reconstruction is trained for full-frame instead of being trained on specific aspects like viseme for speech video reconstruction.\\
%\noindent

Further we introduce a new unit for extracting ROI from the reconstructed frames (as shown in the Figure~\ref{fig:roi_arch}). During the experiments we noticed that the the ROI extraction is performed best from the last three output channels. The ROI extraction unit consists of 2 convolutional layers and a spatial transformer layer proposed by \cite{jaderberg2015spatial}. They key idea is that the convolutional layers learn to focus on the ROI while the STN learns to extract only the mouth region. We show that the FCN3D model shows a much greater boost in terms of metrics for even sequential tasks such as prefix and suffix corruptions, coming close to the performance of that of bi-directional LSTMs for sequential reconstructions. \\
\noindent
The ROI loss can be defined as:

\begin{equation}
	 \mathcal{L}_{roi} = \frac{1}{N} \sum_{x\in X}||f_{stn}(x_{corr}) - y_{roi}||_1 
\end{equation}

where $y_{roi}$ are the ground truth ROI features extracted from the frames and $f_{roi}$ is the ROI stream of the architecture.\\
%\noindent
Thus the total loss during training is 

\begin{equation}
 \mathcal{L}_{total} = \mathcal{L}_{frame} + \mathcal{L}_{ROI}
\end{equation}

%\noindent
However, for computing metrics we take into account only the reconstructed frames since ROI loss is only an auxiliary for frame reconstruction.

%%%%%%%%%%%%%%%%%%%%%%%%%%%%%%%%%%%%%%%%%%%%%%%%%%%%%%%%%%%%%%%%%%%%%%%%

\section{Results and Discussion}\vspace{-1em}
In this section, we present the experimental results for the models given in the Section~\ref{subsec:models} on the tests explained in the Section~\ref{sec:evaluation methods}. Results of the test suite are given in the Table~\ref{tab:metric_table1}. We notice that for random corruption, though fully convolutional models beat bi-directional LSTM for low levels of noise, the bi-directional LSTMs gives a much better performance for longer sequential frame reconstructions. We see that the ROI extraction unit helps boost the performance of the FCN model. The ROI stream network helps attain a better PSNR and SSIM comparable with LSTM. This can be attributed to the model's ability to focus more on visemes indirectly by learning a better reconstruction of the ROI for speech videos. The secondary stream is a very shallow network consisting of two convolutions and a spatial transformer which helps extract the mouth ROI from the already generated output. Thus, ROI loss further helps the network learn more about the visemic structure of the reconstruction and help attain better scores on the metrics.\\
%\noindent

We also show a quantitative comparison with Super SloMo~\cite{jiang2018super}. Since Super SloMo only allows for the prediction of the contiguous frame we had to significantly modify the experimental setup to make any comparisons with it. The architecture did not allow the first and last frame of the batch from being corrupted. Therefore, though the quantitative results are the best for it, Super-Slomo suffers from severe experimental restrictions unlike the other models presented. In the case of continuous corruption, it will require the start and the end frames to reconstruct the intermediate ones, hence it is difficult to implement it in real time while taking into account the random nature of the noise/corruption. 

\par
If we look at the results on viseme and word-level datasets (Table~\ref{tab:metric_table2}), it can be seen that FCN3D with ROI outperforms SuperSlomo in terms of SSIM in word level corruption and PSNR for Viseme corruption attaining a PSNR of 26.54 against 25.2605 by Super SloMo. This is due to the smaller duration of corruption in these datasets as compared to those in Table~\ref{tab:metric_table1}. This is also likely to be the practical corruption distribution for realtime speech video streaming where networks like Super SloMo cannot function since they generate frames for only third time $t_3$ step between $t_1$ and $t_2$ (one of which may not exist). The proposed models are not restricted by this limitation and for the speech videos, capture a longer temporal context (not just the frames at timestamps $t_1$ and $t_2$) besides focusing on the mouth region using the ROI unit. As evident from Table \ref{tab:metric_table1} for random corruption FCN3D+ROI gives a high PSNR value of 24.75 beating BDLSTM and coming in close with Super SloMo with a PSNR value of 28.26, while LSTM outperforms FCN3D+ROI for prefix and suffix corruptions however with a smaller difference of 2.21, while Super SloMo outperforms both with a higher margin due to its ability to incorporate optical flow.
This highlights a key limitation of such methods for speech video interpolation tasks due to their inability to capture context also reinforcing the need for a better test suite for a more faithful evaluation of speech video interpolation.

%%%%%%%%%%%%%%%%%%%%%%%%%%%%%%%%%%%%%%%%%%%%%%%%%%%%5
\section{Conclusion}\vspace{-1em}
In this papaer, we demonstrate the necessity of a comprehensive linguistically-informed test suite that encompasses all the major aspects of speech in the task of speech video interpolation and reconstruction. We release six challenge datasets for this purpose. We also compare several different contemporary deep learning models for the different tasks proposed. We show the importance of incorporating visemic loss and provide a natural proxy to judge the visemic nature of reconstruction. In the future, we would like to cover more such linguistic aspects of speech. A parallel task would also be to take the audio-modality into account while reconstructing the corrupted and missing frames in video interpolation.

%-------------------------------------------------------------------------

\bibliographystyle{IEEEbib}
\bibliography{refs}

\begin{thebibliography}{10}

\bibitem{liu-et-al}
Z.~{Liu}, R.~A. {Yeh}, X.~{Tang}, Y.~{Liu}, and A.~{Agarwala},
\newblock ``Video frame synthesis using deep voxel flow,''
\newblock in {\em 2017 IEEE International Conference on Computer Vision
  (ICCV)}, 2017, pp. 4473--4481.

\bibitem{Meyer15Phase}
Simone Meyer, Oliver Wang, Henning Zimmer, Max Grosse, and Alexander
  Sorkine-Hornung,
\newblock ``Phase-based frame interpolation for video,''
\newblock in {\em Proceedings of the IEEE Conference on Computer Vision and
  Pattern Recognition (CVPR)}, 2015, pp. 1410--1418.

\bibitem{jiang2018super}
Huaizu Jiang, Deqing Sun, Varun Jampani, Ming-Hsuan Yang, Erik Learned-Miller,
  and Jan Kautz,
\newblock ``Super slomo: High quality estimation of multiple intermediate
  frames for video interpolation,''
\newblock in {\em Proceedings of the IEEE Conference on Computer Vision and
  Pattern Recognition}, 2018, pp. 9000--9008.

\bibitem{kumar2017obamanet}
Rithesh Kumar, Jose Sotelo, Kundan Kumar, Alexandre de~Br{\'e}bisson, and
  Yoshua Bengio,
\newblock ``Obamanet: Photo-realistic lip-sync from text,''
\newblock {\em arXiv preprint arXiv:1801.01442}, 2017.

\bibitem{chen2018lip}
Lele Chen, Zhiheng Li, Ross K~Maddox, Zhiyao Duan, and Chenliang Xu,
\newblock ``Lip movements generation at a glance,''
\newblock in {\em Proceedings of the European Conference on Computer Vision
  (ECCV)}, 2018, pp. 520--535.

\bibitem{Guardian-fake-news}
Olivia Solon,
\newblock ``The future of fake news: don't believe everything you read, see or
  hear,''
  url{https://www.theguardian.com/technology/2017/jul/26/fake-news-obama-video-trump-face2face-doctored-content},
  2017.

\bibitem{kumar2019harnessing}
Yaman Kumar, Dhruva Sahrawat, Shubham Maheshwari, Debanjan Mahata, Amanda
  Stent, Yifang Yin, Rajiv~Ratn Shah, and Roger Zimmermann,
\newblock ``Harnessing gans for zero-shot learning of new classes in visual
  speech recognition.,''
\newblock in {\em AAAI}, 2020, pp. 2645--2652.

\bibitem{wang2004image}
Zhou Wang, Alan~C Bovik, Hamid~R Sheikh, and Eero~P Simoncelli,
\newblock ``Image quality assessment: from error visibility to structural
  similarity,''
\newblock {\em IEEE transactions on image processing}, vol. 13, no. 4, pp.
  600--612, 2004.

\bibitem{massaro2014speech}
Dominic~W Massaro and Jeffry~A Simpson,
\newblock {\em Speech perception by ear and eye: A paradigm for psychological
  inquiry},
\newblock Psychology Press, 2014.

\bibitem{chung2016lip}
Joon~Son Chung and Andrew Zisserman,
\newblock ``Lip reading in the wild,''
\newblock in {\em Asian Conference on Computer Vision}. Springer, 2016, pp.
  87--103.

\bibitem{yu2016automatic}
Dong Yu and Li~Deng,
\newblock {\em AUTOMATIC SPEECH RECOGNITION.},
\newblock Springer, 2016.

\bibitem{mikolov2013distributed}
Tomas Mikolov, Ilya Sutskever, Kai Chen, Greg~S Corrado, and Jeff Dean,
\newblock ``Distributed representations of words and phrases and their
  compositionality,''
\newblock in {\em Advances in neural information processing systems}, 2013, pp.
  3111--3119.

\bibitem{wolterink2017deep}
Jelmer~M Wolterink, Anna~M Dinkla, Mark~HF Savenije, Peter~R Seevinck,
  Cornelis~AT van~den Berg, and Ivana I{\v{s}}gum,
\newblock ``Deep mr to ct synthesis using unpaired data,''
\newblock in {\em International workshop on simulation and synthesis in medical
  imaging}. Springer, 2017, pp. 14--23.

\bibitem{rowe1994mpeg}
Lawrence~A Rowe, Ketan~D Mayer-Patel, Brian~C Smith, and Kim Liu,
\newblock ``Mpeg video in software: Representation, transmission, and
  playback,''
\newblock in {\em High-Speed Networking and Multimedia Computing}.
  International Society for Optics and Photonics, 1994, vol. 2188, pp.
  134--144.

\bibitem{mcauliffe2017montreal}
Michael McAuliffe, Michaela Socolof, Sarah Mihuc, Michael Wagner, and Morgan
  Sonderegger,
\newblock ``Montreal forced aligner: Trainable text-speech alignment using
  kaldi.,''
\newblock in {\em Interspeech}, 2017, vol. 2017, pp. 498--502.

\bibitem{afouras2018lrs3}
Triantafyllos Afouras, Joon~Son Chung, and Andrew Zisserman,
\newblock ``Lrs3-ted: a large-scale dataset for visual speech recognition,''
\newblock {\em arXiv preprint arXiv:1809.00496}, 2018.

\bibitem{anina2015ouluvs2}
Iryna Anina, Ziheng Zhou, Guoying Zhao, and Matti Pietik{\"a}inen,
\newblock ``Ouluvs2: A multi-view audiovisual database for non-rigid mouth
  motion analysis,''
\newblock in {\em 2015 11th IEEE International Conference and Workshops on
  Automatic Face and Gesture Recognition (FG)}. IEEE, 2015, vol.~1, pp. 1--5.

\bibitem{yang2019lrw}
Shuang Yang, Yuanhang Zhang, Dalu Feng, Mingmin Yang, Chenhao Wang, Jingyun
  Xiao, Keyu Long, Shiguang Shan, and Xilin Chen,
\newblock ``Lrw-1000: A naturally-distributed large-scale benchmark for lip
  reading in the wild,''
\newblock in {\em 2019 14th IEEE International Conference on Automatic Face \&
  Gesture Recognition (FG 2019)}. IEEE, 2019, pp. 1--8.

\bibitem{xiang2020zooming}
Xiaoyu Xiang, Yapeng Tian, Yulun Zhang, Yun Fu, Jan~P. Allebach, and Chenliang
  Xu,
\newblock ``Zooming slow-mo: Fast and accurate one-stage space-time video
  super-resolution,''
\newblock in {\em IEEE/CVF Conference on Computer Vision and Pattern
  Recognition (CVPR)}, June 2020, pp. 3370--3379.

\bibitem{you2018structurally}
Chenyu You, Qingsong Yang, Hongming Shan, Lars Gjesteby, Guang Li, Shenghong
  Ju, Zhuiyang Zhang, Zhen Zhao, Yi~Zhang, Wenxiang Cong, et~al.,
\newblock ``Structurally-sensitive multi-scale deep neural network for low-dose
  ct denoising,''
\newblock {\em IEEE Access}, vol. 6, pp. 41839--41855, 2018.

\bibitem{guibas2017synthetic}
John~T Guibas, Tejpal~S Virdi, and Peter~S Li,
\newblock ``Synthetic medical images from dual generative adversarial
  networks,''
\newblock {\em arXiv preprint arXiv:1709.01872}, 2017.

\bibitem{vincent2008extracting}
Pascal Vincent, Hugo Larochelle, Yoshua Bengio, and Pierre-Antoine Manzagol,
\newblock ``Extracting and composing robust features with denoising
  autoencoders,''
\newblock in {\em Proceedings of the 25th international conference on Machine
  learning}, 2008, pp. 1096--1103.

\bibitem{48501}
Valentin Bazarevsky, Yury Kartynnik, Andrey Vakunov, Karthik Raveendran, and
  Matthias Grundmann,
\newblock ``Blazeface: Sub-millisecond neural face detection on mobile gpus,''
\newblock 2019.

\bibitem{jaderberg2015spatial}
Max Jaderberg, Karen Simonyan, Andrew Zisserman, et~al.,
\newblock ``Spatial transformer networks,''
\newblock in {\em Advances in neural information processing systems}, 2015, pp.
  2017--2025.

\end{thebibliography}
% \bibliography{biblo}

%-----------------------------------------------------------------------
%%%%%%%%%%%%%%%%%%%%%%%%%%%%%%%%%%%%%%%%%%5

\onecolumn
\section{Appendix}
\begin{table*}[ht]
	\centering
	\begin{tabular}{|c|c|c|c|c|c|}
		\hline
		\\
		Model & Corruption (\%) & MSE & L1 & SSIM & PSNR \\
		\hline
		\hline
		\\
		\textbf{Random Corruption} & & & & & \\
		\hline

		\textbf{FCN3D} & 75 & 0.008719796199 & 0.05576719156 & 0.7754931869 & 20.78947158 \\
		\hline
		\textbf{BDLSTM} & 75 & 0.00585898655 & 0.03845565588 & 0.8575272247 & 23.27513855 \\
		\hline
		\textbf{FCN3D + ROI} & 75 & 0.0042412771 & 0.03334277668 & \textbf{0.8654025606} & \textbf{
		24.75212616}\\
		\hline 
		\textbf{Super SloMo} & 75 & \textbf{0.0032095665} & \textbf{0.01857670849} & \textbf{0.9603758130} & \textbf{28.26607934}\\
		\hline
		\\
		
		% Model & Corruption (\%) & MSE & L1 & SSIM & PSNR \\
		\hline
		\textbf{FCN3D} & 40 & 0.004914738255 & 0.03917299796 & 0.8867508521 & 23.30322587 \\
		\hline
		\textbf{BDLSTM} & 40 & 0.004940670419 & 0.03570744986 & 0.8905155776 & 24.38299802 \\
		\hline
		\textbf{FCN3D + ROI} & 40 & \textbf{0.001789543358} & 0.02319063034 & \textbf{0.9326486659} & \textbf{28.31231905}\\
		\hline
		\textbf{Super SloMo} & 40 & {0.001827087549} & \textbf{0.01178585826} & \textbf{0.9849396967} & \textbf{30.34593392}\\
		\hline
		\\
		\textbf{Prefix Corruption} \\
		\hline
		% Model & Corruption (\%) & MSE & L1 & SSIM & PSNR \\
		\hline
		\textbf{FCN3D} & 75 & 0.03761557525 & 0.1411872019 & 0.4735313937 & 14.27897795 \\
		\hline
		\textbf{BDLSTM} & 75 & 0.01357646521 & 0.06795089568 & \textbf{0.6898999835} & \textbf{19.08365961} \\
		\hline
		\textbf{FCN3D + ROI} & 75 & 0.03076031727 & 0.1183280068 & 0.5208106449 & 15.39350073\\
		\hline
		\textbf{Super SloMo} & 75 & \textbf{0.01147114887} & \textbf{0.0528111567} & \textbf{0.7387903618} & \textbf{20.82188895}\\
		\hline
		\\
		\hline
		% Model & Corruption (\%) & MSE & L1 & SSIM & PSNR \\
		\hline
		\textbf{FCN3D} & 40 & 0.0176491743 & 0.08077385376 & 0.7199462504 & 17.5958132 \\
		\hline
		\textbf{BDLSTM} & 40 & 0.007748413963 & 0.0462029091 & \textbf{0.8265586414} & \textbf{21.73186513} \\
		\hline
		\textbf{FCN3D + ROI} & 40 & 0.01163159572 & 0.05775547531 & 0.7721097014 & 19.67188354\\
		\hline
		\textbf{Super SloMo} & 40 & {0.00629138803} & {0.03642695826} & \textbf{0.8411455717} & \textbf{23.1864927}\\
		\hline
		\\
		\textbf{Suffix Corruption} \\
		\hline
		% Model & Corruption (\%) & MSE & L1 & SSIM & PSNR \\
		\hline
		\textbf{FCN3D} & 75 & 0.036539371 & 0.1388430108 & 0.48635537 & 14.39668367 \\
		\hline
		\textbf{BDLSTM} & 75 & 0.01277490123 & 0.06604159133 & \textbf{0.6951813442} & \textbf{19.21521863} \\
		\hline
		\textbf{FCN3D + ROI} & 75 & 0.03108713292 & 0.1189973814 & 0.5161932313 & 15.34146382\\
		\hline
		
		\textbf{Super SloMo} & 75 & \textbf{0.01606990903} & \textbf{0.0645460370} & \textbf{0.6952581282} & \textbf{19.83196098}\\
		\hline
		% Model & Corruption (\%) & MSE & L1 & SSIM & PSNR \\
 \\		
		\textbf{FCN3D} & 40 & 0.01651665661 & 0.07831238874 & 0.7360948577 & 17.85544448 \\
		\hline
		\textbf{BDLSTM} & 40 & 0.006905819333 & 0.04328332246 & \textbf{0.8464001736} & \textbf{22.30525472} \\
		\hline
		\textbf{FCN3D + ROI} & 40 & 0.01134652429 & 0.05662384959 & 0.777493871 & 19.77901635\\
		\hline
		\textbf{Super SloMo} & 40 & \textbf{0.01155048322} & \textbf{0.04815621729} & \textbf{0.8140604513} & \textbf{22.4334380}\\
		
	\end{tabular}
	\caption{The above table shows the metrics obtained on the test set for the different experiments conducted. For each experiment 2 different levels of corruptions have been tested, 40\% and a very high corruption 75\%. The above table also experimentation highlights how the behaviours of models vary across different experiments, thus further reaffirming the need for a more comprehensive test suite.}
	\label{tab:appendix_metric_table}
\end{table*}

\begin{table}[htbp]
	\centering
	\footnotesize
	\begin{tabular}{|l|l|l|l|}
		\hline \textbf{Metrics} & \textbf{Train} & \textbf{Eval} & \textbf{Test} \\ \hline
		
		\textbf{\# Speakers} & 5091 & 4005 & 413 \\ \hline
		
		\textbf{\# Utterances} & 118,516 & 31,982 & 1321 \\ \hline
		
		\textbf{\# Word instances} & 3,894,800 & 807,375 & 9890 \\ \hline
		
		\textbf{Vocab} & 51,211 & 17,546 & 2002 \\ \hline
		
		\textbf{Avg. Video Length} & XXX & XXX & XXX \\ \hline
	\end{tabular}
	\caption{
		\small
		\label{tab:appendix_LRS3 metrics} Overview of LRS-3 dataset used for experimentation}
\end{table}

\begin{figure*}[!htb]
 \centering
 \begin{tabular}{c }
 \includegraphics[scale=0.35]{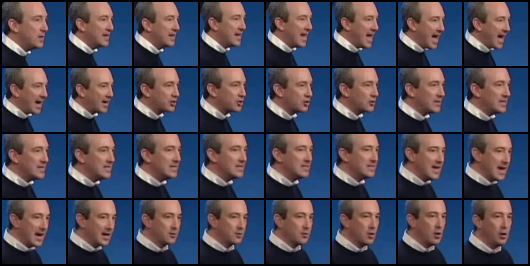} \\
 \includegraphics[scale=0.35]{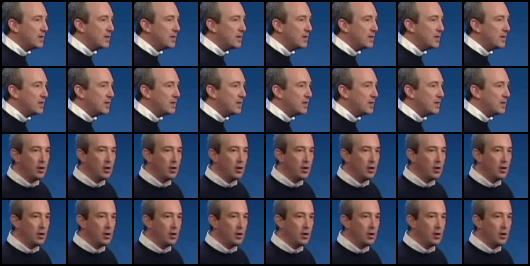}\\
  MSE: \; 0.01086 \\
 PSNR: \; 19.64074 \\  
 SSIM: \; 0.7619643
 \end{tabular}
 \caption{Another example to show the case where we get relatively high SSIM and PSNR values despite the usage of only two frames to create the synthetic clip. This reinforces the fact that we need better metrics to compensate for the underlying context.}
 
 \label{fig:appendix_frame_diff}
\end{figure*}

\begin{table*}
\small
\centering
	\begin{tabular}{|c|c|c|c|c|}
		\hline
% 		\\
		Model & MSE & L1 & SSIM & PSNR \\
		\hline
		\hline
% 		\\
		\multicolumn{5}{|c|}{\textbf{Viseme Corruption}} \\
		\hline
% 		\textbf{FCN3D} & 75 & 0.03294822413 & 0.1267357732 & 0.5301524016 & 14.83449745 \\
		\textbf{FCN3D} & 0.01460618444 & 0.0568077997 & 0.83858460187 & 18.52015423 \\
		\hline
		\textbf{BDLSTM} & 0.02084208424 & 0.0828854019 & 0.72696408065 & 17.28277912 \\
		\hline
		\textbf{FCN3D + ROI} & 0.00229573487 & 0.0268876856 & \textbf{0.91065606555} & \textbf{26.54016680} \\
		\hline 
		\textbf{Super SloMo} & 0.00377557111 & 0.0265828257 & \textbf{0.91293388146} & \textbf{25.26056400} \\
		\hline
		\hline
% 		\\
		\multicolumn{5}{|c|}{\textbf{Word Level Corruption}} \\
% 		\hline
		% Model & Corruption (\%) & MSE & L1 & SSIM & PSNR \\
		\hline
		\textbf{FCN3D} & 0.01476183945 & 0.0596284348 & 0.81098916828 & 18.45544786 \\
		\hline
		\textbf{BDLSTM} & 0.01841032799 & 0.0762009256 & 0.74111890852 & 18.06654904 \\
		\hline
		\textbf{FCN3D + ROI} & 0.00313893227 & 0.0303552080 & 0.88049633604 & 25.32792884 \\
		\hline
		\textbf{Super SloMo} & 0.00426348229 & 0.0304555968 & \textbf{0.88005496244} & \textbf{24.59719328} \\
		\hline
		\hline
% 		\\
		\multicolumn{5}{|c|}{\textbf{Multi Word Corruption}} \\
		\hline
		% Model & Corruption (\%) & MSE & L1 & SSIM & PSNR \\
		\textbf{FCN3D} & 0.01131343627 & 0.0555250485 & 0.80642498433 & 19.61113891 \\
		\hline
		\textbf{BDLSTM} & 0.01841032808 & 0.0762009258 & 0.74111890912 & 18.06654899 \\
		\hline
		\textbf{FCN3D + ROI} & 0.00538372894 & 0.0382826699 & 0.83086033181 & 23.13544090\\
		\hline
		
		\textbf{Super SloMo} & 0.00524661909 & 0.0348772242 & \textbf{0.84817349957} & \textbf{23.82953434}\\
		\hline
		% Model & Corruption (\%) & MSE & L1 & SSIM & PSNR \\
 % \\		
		
	\end{tabular}
	\caption{The above table shows similar metrics for the same models on the newly introduced datasets. It can be seen that simpler models trained with ROI embedding give results close to the baseline.}
	\label{tab:metric_table_viseme_full}
\end{table*}

\twocolumn

\end{document}